# A Data Driven Sequential Learning Framework to Accelerate and Optimize Multi-Objective Manufacturing Decisions

Hamed Khosravi[1], Taofeeq Olajire[2], Ahmed Shoyeb Raihan[3], Imtiaz Ahmed[4, *]

[1]*Department of Industrial & Management Systems Engineering, West Virginia University, Morgantown, WV 26505, ORCID: 0000-0003-0378-6291 Email: hk00024@mix.wvu.edu*

[2]*Department of Industrial & Management Systems Engineering, West Virginia University, Morgantown, WV 26505, ORCID: 0009-0002-6482-6680 Email: too00002@mix.wvu.edu*

[3]*Department of Industrial & Management Systems Engineering, West Virginia University, Morgantown, WV 26505, ORCID: 0009-0005-4016-2666 Email: ar00065@mix.wvu.edu*

[4]*Department of Industrial & Management Systems Engineering, West Virginia University, Morgantown, WV 26505, ORCID: 0000-0003-1577-7384 Email: imtiaz.ahmed@mail.wvu.edu*

[*]*Corresponding Author: imtiaz.ahmed@mail.wvu.edu*

## Abstract

Manufacturing advanced materials and products with a specific property or combination of properties is often warranted. To achieve that it is crucial to find out the optimum recipe or processing conditions that can generate the ideal combination of these properties. Most of the time, a sufficient number of experiments are needed to generate a Pareto front. However, manufacturing experiments are usually costly and even conducting a single experiment can be a time-consuming process. So, it's critical to determine the optimal location for data collection to gain the most comprehensive understanding of the process. Sequential learning is a promising approach to actively learn from the ongoing experiments, iteratively update the underlying optimization routine, and adapt the data collection process on the go. This paper presents a novel data-driven Bayesian optimization framework that utilizes sequential learning to efficiently optimize complex systems with multiple conflicting objectives. Additionally, this paper proposes a novel metric for evaluating multi-objective data-driven optimization approaches. This metric considers both the quality of the Pareto front and the amount of data used to generate it. The proposed framework is particularly beneficial in practical applications where acquiring data can be expensive and resource intensive. To demonstrate the effectiveness of the proposed algorithm and metric, the algorithm is evaluated on a manufacturing dataset. The results indicate that the proposed algorithm can achieve the actual Pareto front while processing significantly less data. It implies that the proposed data-driven framework can lead to similar manufacturing decisions with reduced costs and time.

## 1. Introduction

Over the course of history, the manufacturing industry has been significantly impacted by technological revolutions (Yuan et al., 2022). With ongoing advancements in manufacturing, various technologies such as big data, Industry 4.0, the Internet of Things, cloud computing, and advanced artificial intelligence have emerged (Wang et al., 2021). The use of these technologies provides engineers and manufacturers with exciting new tools to address real-world challenges (Nti et al., 2021). Thanks to the availability of more data and advances in artificial intelligence, smart machines are becoming increasingly intelligent, resulting in more efficient and optimized industries. Despite this, cost and completion time issues are unavoidable in most manufacturing operations (Golab, Massey & Moultrie, 2022). Factors such as production process inefficiencies, supplier issues, and the unpredictable nature of the business can cause cost overruns for many manufacturers (Bandyopadhyay & Traxel, 2018). While some cost problems are inevitable, others can be mitigated through foresight and planning (Ruane, Walsh & Cosgrove, 2023).

In the modern world, decision support tools and systems that promote efficient and effective design, development, and operation of products, processes, and systems are essential for manufacturers (Shao, Brodsky & Miller, 2015). The manufacturing industry generates vast amounts of data daily, but there is still a limited understanding of this field (Davis et al., 2012). In recent times, machine learning algorithms have gained wider usage in manufacturing to leverage available data and create products and items with reduced labor costs, time, and effort requirements (Kumar et al., 2022). The research studies in the area of machine learning in manufacturing have shown that machine learning techniques can be used for a variety of tasks including design, management, scheduling, material resource planning, capacity analysis, quality control, maintenance, and automation (Fahle, Prinz & Kuhlenkötter, 2020). Incorporating machine learning into manufacturing operations presents a number of challenges, including the collection of appropriate data (Kumar et al., 2022). However, collecting new data points can be costly and time-consuming in certain manufacturing applications. One possible solution is to generate large training datasets for every new setup and train a new machine learning model from scratch. However, this approach is not always feasible since future improvements will require generating new data, which can be a difficult task (Ramezankhani et al., 2021). To overcome this issue, ensemble learning and sequential distributed learning approaches have been proposed (Zerka et al., 2021). Sequential learning is particularly advantageous since the model adjusts its performance based on newly acquired training data, leading to optimization (Palizhati et al., 2022). It allows active learning which is more advantageous compared to a passive learning scenario specially in a data scarce environment.

Physical experiments often require optimization of control parameters to achieve the best results, and this has led to increased interest in utilizing machine learning algorithms in this domain (Davletov et al., 2020). Optimization problems can be single-objective or multi-objective, depending on the number of objective functions to minimize or maximize (Rao, Rai & Balic, 2016; Rao, 2010). In the manufacturing industry, various performance criteria, such as surface roughness, dimensional accuracy and tolerance, production cost, mechanical properties, and tribological properties, are considered when evaluating manufacturing processes. It is essential to assess these objectives in relation to the impact of the independent process parameters (Fountas et al., 2020). While single-objective optimization methods have been adopted in manufacturing, researchers have already shown interest in multi-objective optimization (MOO) to design high-quality and low-cost products across all branches of manufacturing engineering (Fountas et al., 2020).

In MOO, a solution may be advantageous in one objective while being detrimental in another, and vice versa (Rao, Rai & Balic, 2016). In MOO problems involving trade-offs between different objectives, a Pareto-optimal solution is generally presented to the decision maker. Such solutions are non-dominant as they cannot increase one objective without decreasing at least one other objective (Ruane, Walsh & Cosgrove, 2023). Several methods are available for tackling MOO problems in manufacturing, with evolutionary algorithms being among the most useful (Jiang, Chen & Liu, 2014; Raju et al., 2018). A wide range of metaheuristic optimization methods, including genetic algorithms and particle swarm optimization, can be employed as well (Balekelayi, Woldesellasse & Tesfamariam, 2022). Additionally, deep learning conditional methods have proven to be highly effective in generating multiple Pareto-optimal solutions. Multi-objective Bayesian optimization has also received significant attention in recent years (Konakovic Lukovic, Tian, & Matusik, 2020).

To evaluate the competing approaches, various performance criteria are commonly used, such as the capacity of the non-dominated solution set, solution convergence to the true Pareto fronts, solution diversity in the objective space, and extent of solution domination relative to reference sets (Okabe, Jin & Sendhoff, 2003). Nonetheless, there is still an opportunity to develop novel MOO metrics suitable for concave Pareto fronts in hypervolume calculations (Zitzler & Thiele, 1999) (Jiang et al., 2014).

The manufacturing industry is currently encountering a significant obstacle in achieving economic goals, prompting researchers to explore optimization techniques as potential solutions (Rao, Rai & Balic, 2016). However, many existing optimization methods rely on functional relationships, which may not accurately reflect real-world scenarios where data points without known relationships, also known as black-box functions, are more common. Furthermore, conventional prediction and optimization approaches are highly reliant on the quantity of available data points during the training process and display diminished performance with smaller datasets. It is important to note that data points are considered a luxury for the manufacturing industry, necessitating the development of approaches that use them prudently during the optimization process. This study aims to tackle these issues and provide the following primary contributions.

- We propose a novel data-driven multi-objective Bayesian Optimization framework that can sequentially model the underlying complex system using a Gaussian Process (GP)-based surrogate model.
- We use qParEGO (Daulton, Balandat & Bakshy, 2020) based acquisition function to actively select experiment location in each iteration and consequently update and adjust the optimization model. The outline of this proposed method is depicted in Figure 1.
- We introduce a new performance criterion for assessing models in multi-objective optimization problems. This criterion considers both the Proportional hypervolume (PHV) (Aboutaleb et al., 2017) and the ratio of data utilization, which is particularly critical in real world applications where data collection is costly and time-consuming.
- To simulate the active learning environment, we design and demonstrate our algorithm using a real manufacturing dataset considering two different scenarios. In both cases, our algorithm sequentially unwraps the underlying functional relationship hidden in the data using only a small subset of the available data.

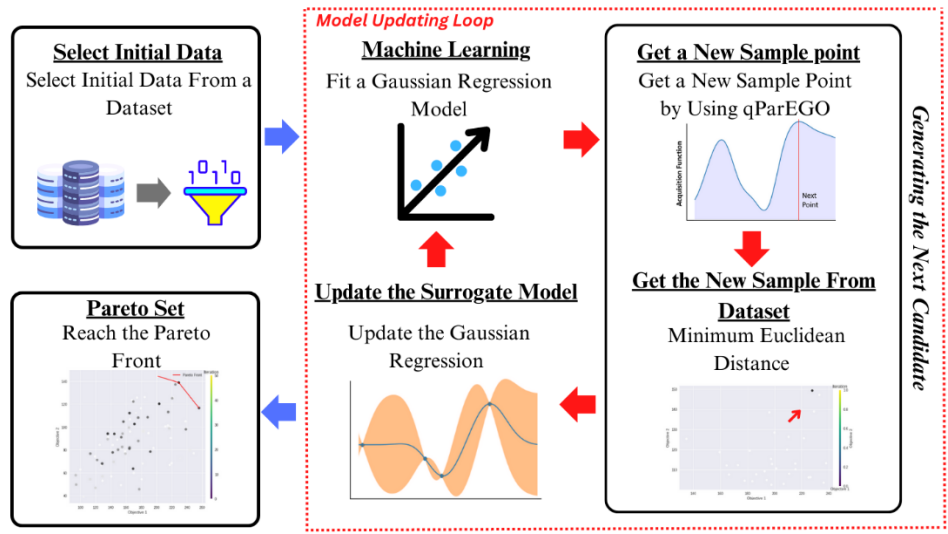

**Fig. 1**: The overview of the proposed method

The rest of the paper is organized as follows. The single and multi-objective optimization approaches commonly utilized in manufacturing industries are summarized in Section 2. Section 3 presents the manufacturing dataset used in this study. Section 4 describes the preliminaries, definitions, steps of our proposed method and performance criterion. Section 5 highlights simulation, computational results, and the comparative study. Finally, Section 6 provides concluding remarks and suggestions for future research.

## 2. Literature Review

This section utilizes an integrated method to review the pertinent literature and gather the requisite information. The review process stages are illustrated in Figure 2. To address the research questions, a comprehensive analysis of published literature in the English language from the Scopus database is performed using both bibliometric and systematic methods.

### 2.1. Comprehensive Review of Relevant Literature

Optimization involves finding the most efficient solution to a given problem. The evaluation of a solution relies on various criteria, which are influenced by the engineer's or researcher's experience, perspective, practical judgment, and the nature of the problem. For example, one might aim to reduce overall costs while maintaining desired performance (Golab, Massey & Moultrie, 2022). The increasing reliance on online services and digital devices has resulted in a massive influx of data (Bandyopadhyay & Traxel, 2018). Industries often collect and analyze these data to demonstrate their potential for future performance. Moreover, these data can be used to train machines to perform specific tasks and further processing through the development of predictive models (Ruane, Walsh & Cosgrove, 2023). In recent years, there has been an increase in research on the application of these generated data on optimization in the manufacturing industry. In recent years, there has been an increase in research on the application of these generated data on optimization in the manufacturing industry. Based on the approach and results of previous studies, Table 1 provides a comprehensive overview of studies on single and multi-objective optimization methods in manufacturing.

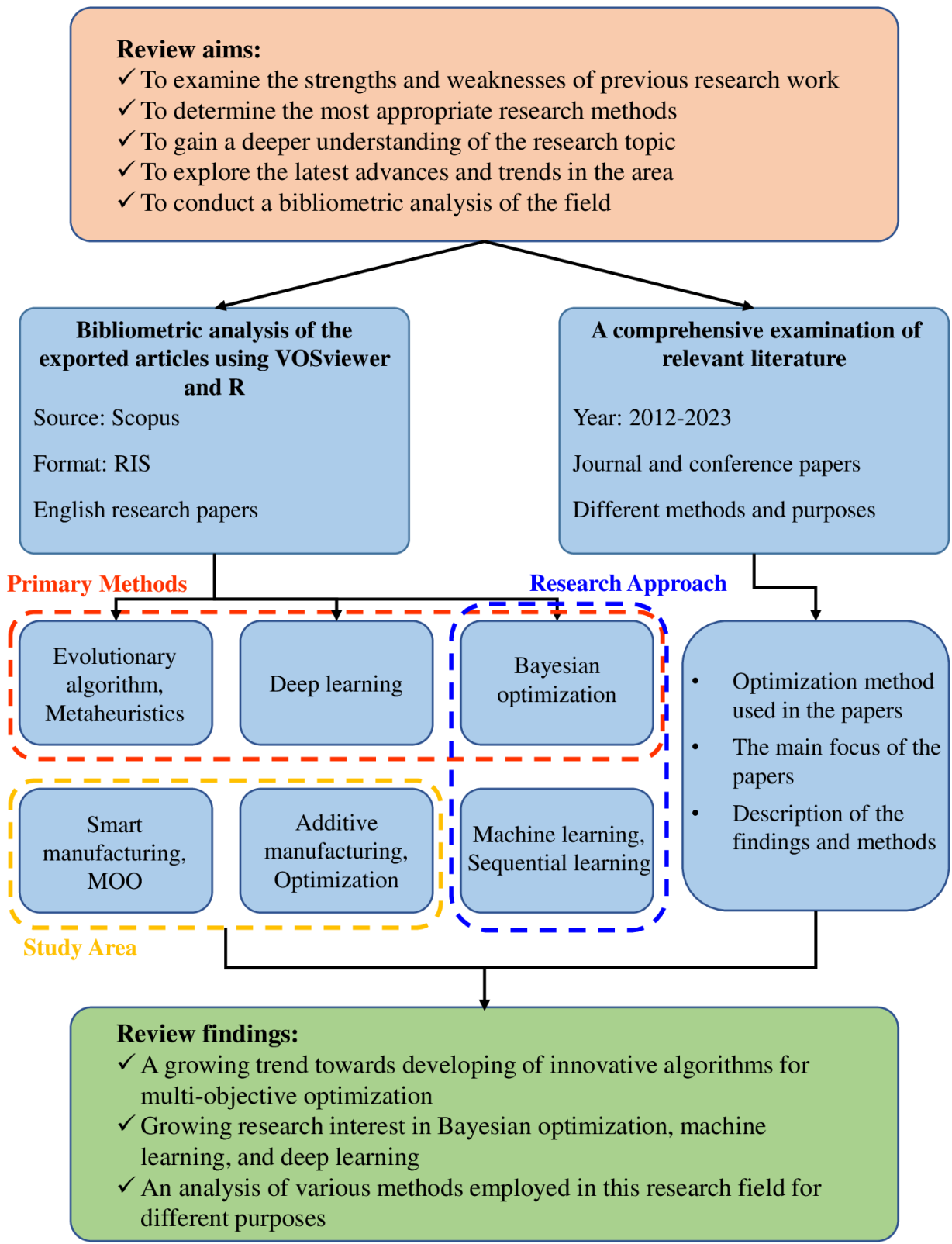

**Fig. 2:** The flowchart of the review methodology

Table 1: A Summary of prior studies regarding optimization in the manufacturing industry

| Ref. | Year | Optimization method used | Focus |
|---|---|---|---|
| (Al Hazza et al., 2012) | 2012 | Multi-objective genetic algorithm (MOGA) | Minimize power consumption cost |
| (Iqbal et al., 2013) | 2013 | Fuzzy rule-based system | Energy consumption, tooling costs, and productivity |
| (Kumar & Singh, 2014) | 2014 | Taguchi 's approach and utility concept | Axial force, radial force, main cutting force and material removal rate |
| (Kübler, Böhner & Steinhilper, 2015) | 2015 | Meta-heuristic genetic algorithm | Resource consumption, machining time, and machining cost |
| (Li et al., 2015) | 2015 | Multi-objective optimization based on neural network | Machining time, energy consumption, and surface roughness |
| (Alizadeh Afrouzy et al., 2016) | 2016 | Fuzzy stochastic multi-objective model | Profit of the supply chain, customer satisfaction, and production of the new products |
| (Lin et al., 2017) | 2017 | Multi-objective teaching–learning-based optimization algorithm | Carbon emissions, operation time, and machining cost |
| (Solomou et al., 2018) | 2018 | Bayesian Optimal Experimental Design (BOED) | Multi-objective materials discovery problems |
| (Asadollahi-Yazdi, Gardan & Lafon, 2018) | 2018 | Non-Dominated Sorting Genetic Algorithm-II (NSGA-II) | Production time and material mass |
| (Dong, Marleau-Finley & Zhao, 2019) | 2019 | Financial Portfolio Approach to the Product | Hybrid artificial intelligence and robust optimization |
| (Ramírez-Márquez et al., 2019) | 2019 | Hybrid algorithm called differential evolution with Taboo List (DETL) | Economics, environmental, and safety (Production of Solar-Grade Silicon) |
| (Greinacher et al., 2020) | 2020 | Meta-model-based multi-objective optimization | Lean and resource efficient manufacturing systems |
| (Fountas et al., 2020) | 2020 | Different metaheuristic algorithms | Fused deposition modelling |

| Ref. | Year | Optimization method used | Focus |
|---|---|---|---|
| (Simon, 2020) | 2020 | A developed multi-objective optimization model | Tooth contact stresses, angular displacement error of the driven gear, and energy losses |
| (Cao et al., 2021) | 2021 | Design of experiment + Meta-model construction + WOA optimization | Surface roughness and dimensional accuracy |
| (Karlsson, Bandaru & Ng, 2021) | 2021 | Combination of NSGA-II and Flexible Pattern Mining ( FPM-NSGA-II) | Considering the preferences of the decision maker |
| (Daulton et al., 2022) | 2022 | Robust Multi-Objective Bayesian Optimization | Optimizing the multivariate value-at-risk (MVaR) |
| (Khatamsaz et al., 2022) | 2022 | A novel multi-information BO framework | Materials design by optimizing multiple objectives |
| (Hu et al., 2023) | 2023 | Multi-objective Bayesian optimization algorithm guided finite element method (FEM) | Optimization of Triply periodic minimal surface (TPMS) structure |
| (Chepiga et al., 2023) | 2023 | Efficient Multi-Objective Bayesian Optimization Algorithm | Parameter optimization for the laser powder bed fusion process |
| (Geng et al., 2023) | 2023 | A multi-objective Bayesian optimization (BO) framework | Parameter settings of physical design tools |
| (Duquesnoy et al., 2023) | 2023 | Machine learning-assisted multi-objective optimization method | High computational cost of manufacturing parameters |

Table 1 presents various studies on multi-objective optimization techniques. Hazza et al. (2012) utilized a multi-objective genetic algorithm to minimize power consumption costs while considering various constraints. Iqbal et al. (2013) proposed a fuzzy rule-based system for multi-objective optimization.

Several studies have investigated optimization models and techniques to improve manufacturing processes. Kumar & Singh (2014) proposed a simplified multi-characteristic optimization model based on Taguchi's design approach. The study optimized four performance characteristics using the utility concept. In a subsequent study, Kübler, Böhner & Steinhilper (2015) used an algorithm based on GA-based non-dominated sorting-II to determine the Pareto front for multi-objective optimization. They conducted a posteriori multi-objective optimization of resource consumption for a multi-pass turning process, examining the algorithm's capability. Another study by Li et al. (2015) presented a multi-objective optimization model for evaluating and determining the optimal cutting parameters in a cutting scheme. The proposed model was shown to align more closely with the characteristics of engraving and milling processes and effectively address the difficulties of machining sculptured parts.

Afrouzy et al. (2016) designed a multi-objective, multi-period, multi-product supply chain model. They presented the proposed model as a single-objective mixed integer programming model using the updated multi-choice goal programming approach. In a subsequent study, Lin et al. (2017) aimed to optimize metal-cutting parameters during multi-pass turning operations. They found that cutting fluids had a substantial impact on reducing carbon emissions, machining costs, and enhancing production efficiency. In 2018, Solomou et al. proposed a Bayesian optimal experimental design called BOED to efficiently discover targeted NiTi shape memory alloys. Their study aimed to propose and apply a closed-loop Bayesian Optimization and Experimental Design (BOED) framework to optimize the search for precipitated Shape Memory Alloys (SMAs) with desired properties that fulfill multiple objectives. Another study conducted in the same year (Asadollahi-Yazdi, Gardan & Lafon, 2018) proposed an approach to select the best manufacturing parameters for fabrication with Fused Deposition Modeling (FDM).

A novel method for calculating the risk of a product portfolio, utilizing an improved artificial intelligence and robust optimization technique was introduced in 2019 (Dong, Marleau-Finley & Zhao, 2019). The study employed a hybrid meta-heuristic algorithm to predict future demand for each product type, followed by the development of a new methodology for calculating product risk and two optimization models. In the same year, a study (Ramírez-Márquez et al., 2019) was performed on the safety and environmental impact of industrial processes as design criteria, using a multi-objective optimization method with the DETL approach to determine optimal parameters for each process. The study also utilized the Pareto Front to determine the best values for each objective function.

In 2020, several studies focused on enhancing manufacturing processes through the application of optimization techniques. Greinacher et al. (2020) proposed a multi-objective optimization method to improve lean and resource-efficient manufacturing processes. They developed a simulation framework to calculate meta-models and perform optimization using meta-models. Fountas et al. (2020) examined the application of various metaheuristic algorithms on single and multi-objective optimization problems related to fused deposition modeling (FDM) in additive manufacturing. These problems involved mechanical tests to evaluate material properties such as tensile strength, compressive strength, and flexural strength. Additionally, Simon (2020) presented a novel multi-optimization method to enhance the performance of hypoid gear pairs. This method utilized fifth-order optimized polynomial functions to vary machine tool settings during the machining process of tooth surfaces, which were then transferred to the numerical controlled machine tool for hypoid gear manufacture using a new algorithm.

A study conducted in 2021 introduced a data-driven approach to determine the most optimal Laser powder bed fusion process parameters that can yield satisfactory surface roughness and dimension accuracy (Cao et al., 2021). The study examined the impact of crucial process parameters on surface roughness and dimensional accuracy of final products. To reflect the final product's dimensional accuracy and surface roughness under varying process parameters, the researchers developed a machine learning model. The Whale Optimization Algorithm (WOA) then used the predicted values as objectives to find the optimal process parameters. In a separate study, Karlsson, Bandaru & Ng (2021) presented the FPM-NSGA-II algorithm, which aimed to not only generate a set of preferred solutions but also describe them using explicit and easily interpretable rules. The researchers combined the NSGA-II algorithm with a data mining approach, called Flexible Pattern Mining (FPM), which could extract knowledge as rules in a continuous and automated fashion. This integration was designed to guide the optimization process towards a solution space that the decision maker would prefer in the objectives space.

In 2022, a study by Daulton et al. introduced the first multi-objective Bayesian optimization (BO) method that is robust to input noise (Daulton et al., 2022). The authors defined MO robust optimization as the optimization of the global set of multiple variable outputs under noisy input. Empirical demonstrations of robust MOBO on real-world problems were also conducted. To efficiently approximate Pareto-optimal parameter configurations, an information gain-based acquisition function was used to sequentially select candidates for tool simulation. Another study in 2022 developed an innovative framework for active learning of materials design with multiple objectives and constraints using multi-information Bayesian Optimization (BO) (Khatamsaz et al., 2022). The framework was demonstrated by exploring the Refractory Multi-Principal-Element Alloy (MPEA) space and optimizing the alloys in the BO Pareto-front. The performance of the alloys was evaluated by the framework and analyzed using density-functional theory to gain an understanding of their superior performance.

A study conducted in 2023 aimed to expedite the design of Titania TPMS architectures by developing a BO-guided FEM analysis procedure (Hu et al., 2023). This was followed by a numerical simulation to obtain the necessary data for finite element analysis. Additionally, a new MOBO algorithm optimized the L-PBF process, identifying three Pareto front points in six iterations and significantly reducing experiment numbers (Chepiga et al., 2023). Furthermore, Geng et al. presented a parameter tuning flow for a physical design tool using an information gain-based multi-objective BO framework surrogated with a multi-task Gaussian process model (Geng et al., 2023). Duquesnoy et al. proposed a standardized methodology utilizing deterministic machine learning to optimize the multi-objective optimization of Lithium-Ion Battery electrode properties and inverse design of the manufacturing process (Duquesnoy et al., 2023).

However, there remains a need for further research in the area of process optimization for high-quality and cost-efficient manufacturing (Jin et al., 2023). To gain deeper insights into current trends in this field, a bibliometric analysis is conducted and described in the following section.

## 2.2. Bibliometric Analysis of the Literature

This section describes the bibliometric analyses that were conducted on a collection of 2361 research papers, obtained in RIS and BibTeX format from Scopus, using keywords relevant to the area of study. The analyses were performed using VOSviewer software (van Eck & Waltman, 2009) and Bibliometrix (Aria & Cuccurullo, 2017). Initially, a network visualization was created to depict the authors' keywords, with a minimum threshold for keyword occurrences set to 30, resulting in 127 keywords, as shown in the map visualization.

Figure 3 displays an overlay map that has been adjusted based on the frequency of occurrences, indicating that research papers in this field can be divided into five clear clusters. The largest cluster, Cluster 1, encompasses a substantial number of keywords such as Bayesian networks, Bayesian optimization, artificial intelligence, machine learning, and Gaussian processes, and is where the current research topic belongs. In other words, the research in question falls within Cluster 1, which is distinguished by the significant occurrence of these keywords. To analyze the progression of topics in this field over time, Figures 4 and 5 present the trending topics and thematic evolution. Figure 4 clearly indicates that machine learning is the dominant trend in this field of study, followed closely by the trend of mechanical properties. It is noteworthy that the present research investigates both of these leading trend topics, which have significant implications for the advancement of knowledge and the development of practical applications in the field. Figure 5 provides a clear overview of the focus of research in this field over time. Prior to 2007, research in this area was primarily focused on scheduling, with a notable emphasis on the design of experiments. However, over time, the trend has shifted towards the use of Bayesian optimization techniques, with recent research focusing on additive manufacturing and robust design, which are closely related to Bayesian optimization. As a result of these trends, advanced optimization techniques are becoming increasingly important in this field as researchers strive to design and manufacture materials with desirable mechanical properties more efficiently and effectively. Specifically, according to data exported from Scopus, there is a rising trend in the number of documents published annually on Bayesian optimization in the manufacturing industry, which is shown in Figure 6.

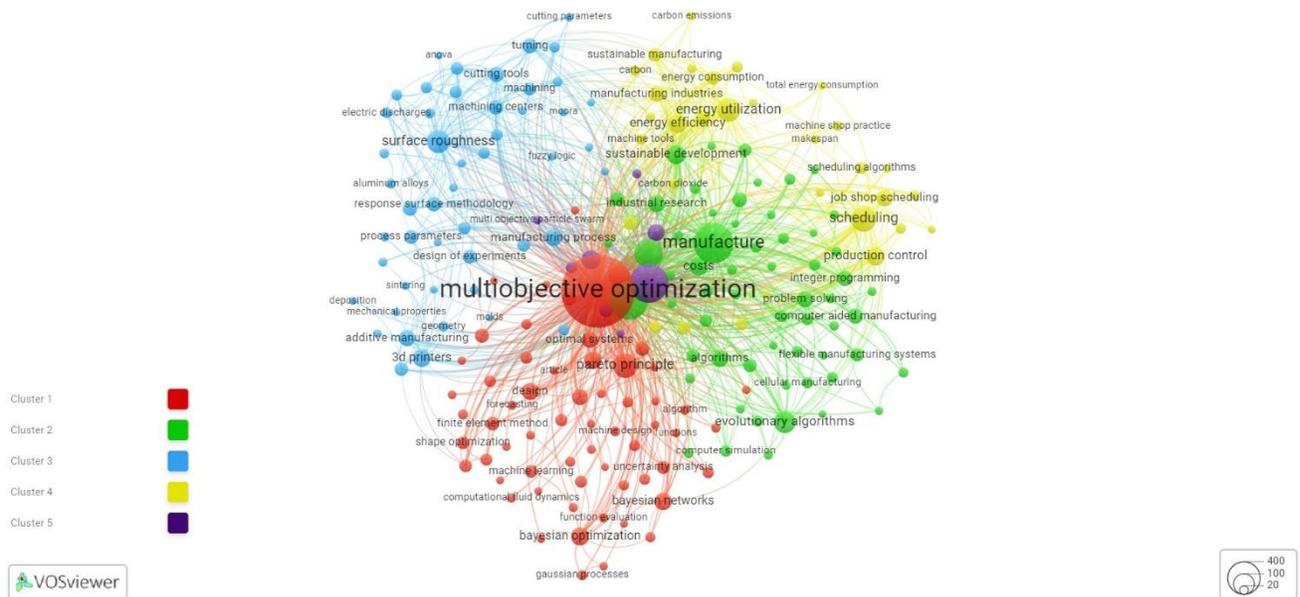

**Fig 3**: Network visualization map, analysis of co-occurrence of author keywords.

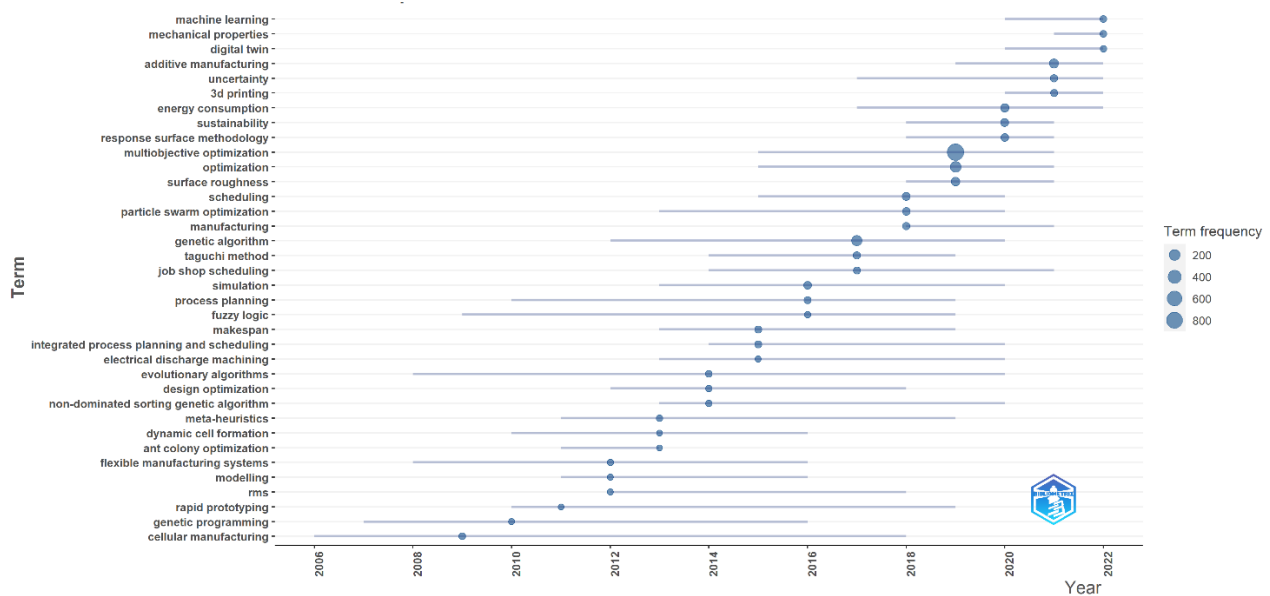

**Fig 4**: Trend topics in the field of multi-objective optimization in manufacturing

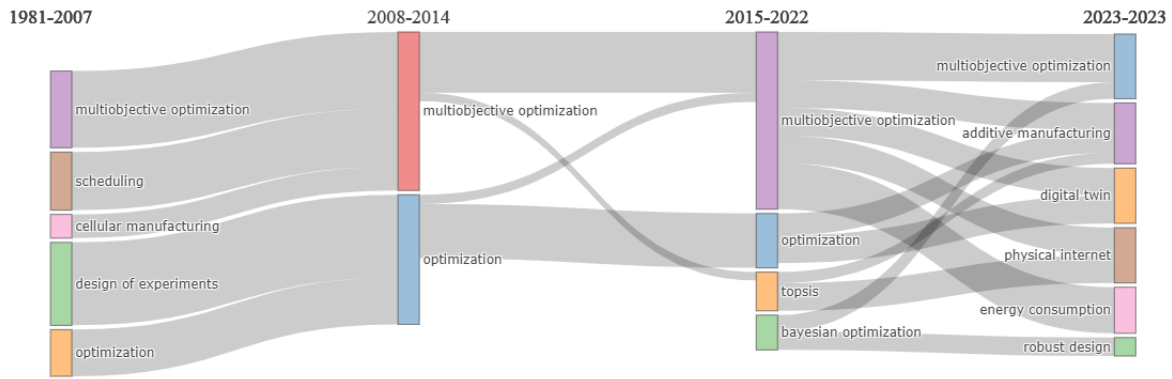

**Fig 5**: Thematic evolution of the keywords in the field of multi-objective optimization on manufacturing

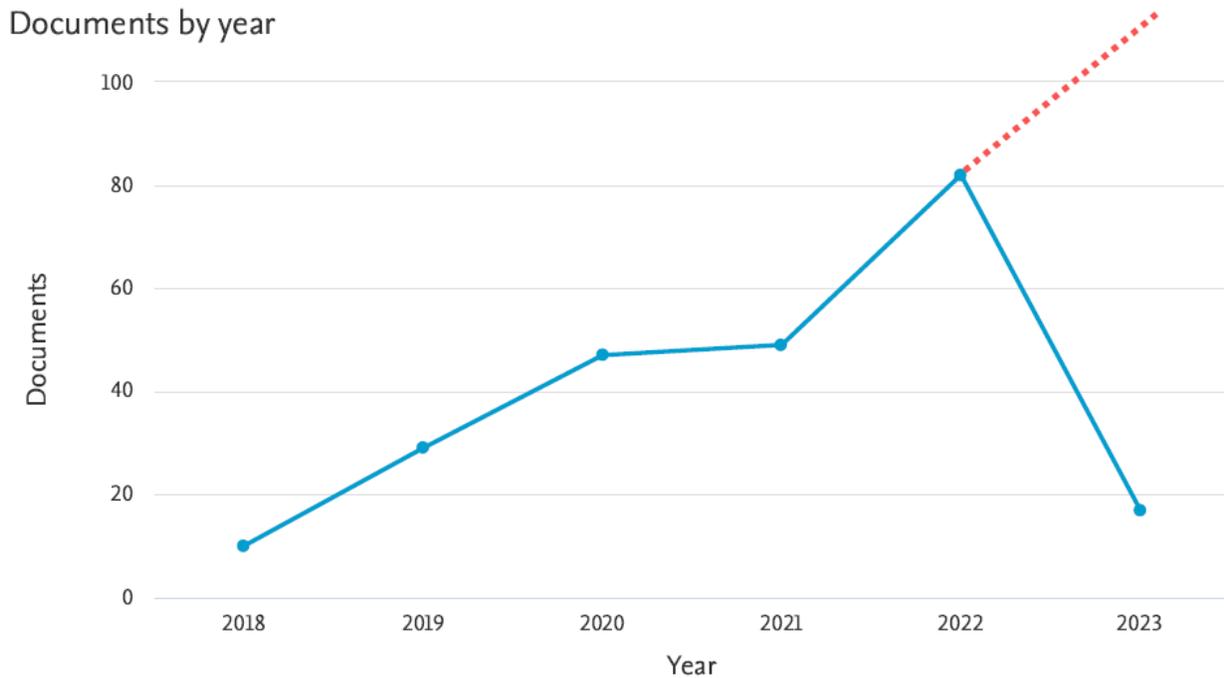

**Fig 6**: The number of publications related to "Bayesian Optimization + Manufacturing" in recent years.

According to Figure 6, the number of published documents in this field has been steadily increasing in recent years. Based on this trend, it is reasonable to assert that the number of documents published in this field will exceed 100 in 2023.

## 2.3. Review of Performance Metrics

In recent years, there has been a surge in the development of new algorithms for MOO problems, and numerous performance metrics have been created to evaluate the efficacy of these algorithms in approximating Pareto fronts (Audet et al., 2021). An analysis of existing literature indicates that the HV indicator is the most commonly utilized performance metric (Riquelme, Von Lucken & Baran, 2015). Future research could focus on identifying alternative performance metrics that overcome the limitations of the HV indicator (such as its exponential cost) while preserving its favorable characteristics (Audet et al., 2021). Table 2 offers a comprehensive comparison of the primary and most significant performance metrics employed in MOO techniques. The comparison is based on several aspects, including convergence, distribution, algorithm comparison, stopping criteria, adaptivity, and data utilization ratio.

| Metric | Convergence | Comparison of algorithms | Distribution | Stopping criteria | Data utilization ratio | Adaptivity |
|---|---|---|---|---|---|---|
| HV | ■ | ■ | ■ | ■ | - | - |
| PHV | ■ | ■ | ■ | ■ | - | - |
| Generational Distance (Van Veldhuizen, 1999) | ■ | ■ | - | - | - | - |
| Epsilon-indicator (Zitzler et al., 2003) | - | ■ | - | ■ | - | - |
| **Adjusted Proportional Hypervolume (Proposed metric in this study)** | ■ | ■ | ■ | ■ | ■ | ■ |

Table 2: A summary of the most important performance metrics for multi-objective optimization

Table 2 highlights the primary advancements and benefits of Adjusted Proportional Hypervolume (APHV), which includes the incorporation of data quantity used to generate the Pareto front and its adaptability to address specific problems and objectives. This novel metric improves the accuracy and efficacy of the evaluation process, making it more robust and efficient in a variety of contexts.

## 3. Data Understanding

In this study, we utilize our multi-objective Bayesian optimization model framework to investigate the MAX ternary carbide/nitride space through Density Functional Theory (DFT) calculations. The MAX phases, or Mn+1AXn, are hexagonal, layered early transition metal carbides and nitrides. M represents a transition metal, A denotes group IV and VA elements in the periodic table, and X represents either Carbon or Nitrogen (Barsoum, 2013). The presence of both metallic and metallic/covalent bonds in the layered structures of the MAX phases produces a range of properties that fall between those of ceramics and metals (Barsoum, 2000; Barsoum & Radovic, 2011; Sun, 2011; Radovic & Barsoum, 2013). Despite the limited number of synthesized pure ternary MAX phase compositions, there is a significant opportunity to discover promising chemistries with optimal property sets by considering various stacking sequences and deviations from stoichiometric ratios at the M, A, and X sites (Amini & Barsoum, 2010; Arróyave et al., 2016; Talapatra et al., 2016).

Due to their wide range of property values and diverse chemistry, MAX phases are an appropriate material system for testing simulation-driven frameworks for materials discovery (Aryal et al., 2014). Our model framework focuses on the MAX phases with M2AX and M3AX2 stoichiometries in their material design space (MDS). The MDS comprises a total of 402 MAX phases, and our objectives are: first, to identify the materials with the maximum bulk modulus (K-exp) and shear modulus (G-exp), and second, to identify materials with the maximum bulk modulus and minimum shear modulus. Maximizing both the bulk modulus and shear modulus can result in a material with high stiffness, meanwhile, materials with high values of bulk modulus and low shear modulus are typically desirable for applications that require strength, and resistance to deformation.

As with most other sequential learning approaches, our work assumes prior knowledge of the MDS of the MAX phases, which is available before conducting the sequential experiments. We represent this prior knowledge as a set of input features that have relationships with the desired properties we are seeking. Talapatra et al. (2018) used their domain knowledge and a thorough literature review to consider a list of features to represent each MAX phase candidate. The table below outlines all the initial features and their descriptions that impact the material properties of the MAX phase candidates.

Table 3: Input features with description

| Feature | Description |
|---|---|
| $C$ | Empirical constants. |
| $m$ | |
| $C_v$ | Valence electron concentration |
| $\frac{e}{a}$ | Electron-to-atom ratio |
| $a$ | Lattice parameters |
| $c$ | |
| $Z$ | Atomic number |
| $Ductility$ | Material's ability to deform under stress without fracturing |
| $Type$ | The specific type of MAX phase material |
| $I_{dist}$ | Interatomic distance |
| $Col_M$ | The groups according to the periodic table of the M, A & X elements |
| $Col_A$ | |
| $Col_X$ | |
| $APF$ | Atomic packing factor |
| $rad$ | Average atomic radius |
| $vol$ | Volume/atom |

## 4. Methodology

This section provides a brief explanation of the methods and terms used in this paper, followed by a discussion of our proposed method and metric.

## 4.1. Multi-objective Optimization

The problem of multi-objective optimization can be formulated as below (Alvarado-Iniesta et al., 2018):

$$\min(or \max) \ F(\mathbf{x}) = [f_1(\mathbf{x}), f_2(\mathbf{x}), \dots, f_m(\mathbf{x})]$$

subject to

$$g(\mathbf{x}) = [g_1(\mathbf{x}), g_2(\mathbf{x}), \dots, g_p(\mathbf{x})] \leq 0$$

$$h(\mathbf{x}) = [h_1(\mathbf{x}), h_2(\mathbf{x}), \dots, h_p(\mathbf{x})] = 0 \tag{1}$$

where $\mathbf{x} = [x_1, x_2, \dots, x_n]$ is a vector of decision variables, $f_i(\mathbf{x})$ is the $i$-$th$ objective function, and $g_j(\mathbf{x})$ and $h_j(\mathbf{x})$ are the inequality and equality constraints, respectively. In most cases, it is not feasible to find a singular solution $\mathbf{x}$ that maximizes (or minimizes) all $m$ objectives while also satisfying all the constraints. Therefore, to evaluate objective vectors, it is customary to use Pareto domination.

## 4.2. Pareto Front

A non-dominated solution $\mathbf{x}^*$ is Pareto optimal if there does not exist another solution $\mathbf{x}'$ in $S$ (let $S$ be the solution space) such that $f_i(\mathbf{x}') \leq f_i(\mathbf{x}^*)$ for all $i = 1, 2, \dots, m$ and $f_i(\mathbf{x}') < f_i(\mathbf{x}^*)$ for at least one $i$ (Natarajan et al., 2019). The set of all Pareto optimal solutions is called the Pareto front, denoted as PF. Therefore, mathematically, PF can be expressed as:

$$PF = \{\mathbf{x}^* \in S \mid \nexists \mathbf{x}' \in S, \mathbf{x}' \neq \mathbf{x}^*, f_i(\mathbf{x}') \leq f_i(\mathbf{x}^*) \forall i = 1, 2, \dots, m \land f_i(\mathbf{x}') < f_i(\mathbf{x}^*) \exists j\} \tag{2}$$

## 4.3. Bayesian Optimization

In general, Bayesian optimization can be summarized as follows:
- Defining the objective function $f(\mathbf{x})$ (typically modeled with a Gaussian process (GP)) requires defining a prior distribution. This distribution can reflect any prior knowledge or assumptions about the function and can be updated over time as new information becomes available.
- Using the prior, an acquisition function is constructed to measure the likelihood of improvement at each point $\mathbf{x}$ in the search space. Common acquisition functions include Expected Improvement (EI), Probability of Improvement, and Upper Confidence Bound.
- While selecting the next experimental point $\mathbf{x}$, the acquisition function tries to balance the trade-off between exploring new regions of the search space and exploiting regions that contain the local optimum.
- Conducting the experiment at the suggested location and adding the new data points to the observed values.
- With the arrival of each new data, the prior distribution is updated to a posterior distribution which reflects the new understanding of the underlying system.
- Repeat step 2-5 until a satisfactory solution is reached.

## 4.4. Gaussian Process

The underlying complex system or design system is modeled using a surrogate model known as the Gaussian process (GP). It captures the beliefs about the relationships between input variables and output variables. The GP model provides estimates of both the predicted mean $\mu_t(\mathbf{x})$ and the epistemic uncertainty $\sigma_t(\mathbf{x})$ at any point $\mathbf{x}$ and time t within the input space. These estimates are based on a given set of observations $\mathcal{D}_{1:t} = \{(\mathbf{x_1}, y_1), (\mathbf{x_2}, y_2), \dots, (\mathbf{x_t}, y_t)\}$, where $\mathbf{x_t}$ refers to the input variables of a process, while $y_t$ refers to the corresponding output variable at time t (Greenhill et al., 2020). In the multi-objective problem, we have more than one $y_t$.

A GP can be fully described by its mean function $m(\mathbf{x})$ and covariance function $k(\mathbf{x}, \mathbf{x}')$:

$$f(\mathbf{x}) \sim \mathcal{GP}\big(m(\mathbf{x}), k(\mathbf{x}, \mathbf{x}')\big) \tag{3}$$

Often, this function $k(\mathbf{x}, \mathbf{x}')$ is referred to as the "kernel", because it measures the smoothness of the process. It is expected that if two points $\mathbf{x}$ and $\mathbf{x}'$ are close, then their corresponding process outputs $y$ and $y'$ will be close as well, with the closeness depending on the distance between the points. The squared exponential (SE) function, also known as the radial basis function (RBF), is one of the most popular choices for the covariance function (Greenhill et al., 2020):

$$k(\mathbf{x}, \mathbf{x}') = \exp\left(-\frac{1}{2\theta^2}\|\mathbf{x} - \mathbf{x}'\|^2\right) \tag{4}$$

An experimental setting involves including a term for normally distributed noise $\varepsilon \sim \mathcal{N}(0, \sigma_{\text{noise}}^2)$, and the observation model can be summarized as follows:

$$y = f(\mathbf{x}) + \varepsilon \tag{5}$$

With Gaussian process regression, it is possible to predict the value of the objective function $f(\cdot)$ at time $t + 1$ for any location $\mathbf{x}$. This results in a normal distribution with mean $\mu_t(\mathbf{x})$ and uncertainty $\sigma_t(\mathbf{x})$ (Greenhill et al., 2020). The posterior distribution (equation 6) reflects the current understanding of the design space and can be sequentially updated with each new experiment (Ahmed et al., 2021).

$$P(f_{t+1} \mid \mathcal{D}_{1:t}, \mathbf{x}) = \mathcal{N}\big(\mu_t(\mathbf{x}), \sigma_t^2(\mathbf{x})\big) \tag{6}$$

where

$$\mu_t(\mathbf{x}) = \mathbf{k}^T \big[\mathbf{K} + \sigma_{\text{noise}}^2 \mathbf{I}\big]^{-1} y_{1:t}$$

$$\sigma_t(\mathbf{x}) = k(\mathbf{x}, \mathbf{x}) - \mathbf{k}^T \big[\mathbf{K} + \sigma_{\text{noise}}^2 \mathbf{I}\big]^{-1} \mathbf{k} \tag{7}$$

$$\mathbf{k} = [k(\mathbf{x}, \mathbf{x_1}), k(\mathbf{x}, \mathbf{x_2}), \dots, k(\mathbf{x}, \mathbf{x_t})]$$

$$\mathbf{K} = \begin{bmatrix} k(\mathbf{x_1}, \mathbf{x_1}) & \dots & k(\mathbf{x_1}, \mathbf{x_t}) \\ \vdots & \ddots & \vdots \\ k(\mathbf{x_t}, \mathbf{x_1}) & \dots & k(\mathbf{x_t}, \mathbf{x_t}) \end{bmatrix} \tag{8}$$

### 4.5. Acquisition Function (qParEGO)

This is an extension of ParEGO that incorporates parallelism, allowing us to jointly optimize and determine the next best location, $x_{best,k}$, in a batch setting where $X_{best,k} = \{x_{best,k,1}, x_{best,k,2}, \dots, x_{best,k,q}\}$, $k$ denotes the iteration number, and $q$ represents the batch size (Biswas et al., 2021). ParEGO is an advanced optimization algorithm that uses a global model to solve complex optimization problems involving multiple objectives (Knowles, 2006). It is modified from the efficient global optimization (EGO) algorithm, which is designed for optimizing a single objective (Jones, Schonlau & Welch, 1998). Global model-based algorithms rely on the use of surrogate models at each iteration to guide the optimization of the black-box function. These surrogate model-based approaches are also referred to as the design and analysis of computer experiments (DACE) (Simpson et al., 1998). The main idea of EGO is to use a DACE model F as a surrogate of the objective function f and calculate an improvement criterion I (x) according to equation 9 (Davins-Valldaura et al., 2017).

$$I\ (\mathbf{x}) = \begin{cases} f_{min} - \hat{f}(\mathbf{x}) & if\ \hat{f}(\mathbf{x}) < f_{min} \\ 0 & otherwise \end{cases} \tag{9}$$

The equation mentioned above includes two terms: $f_{min}$, which is the lowest evaluation obtained by $f$, and $\hat{f}(\mathbf{x})$, which represents the predicted value of the surrogate model F. The surrogate model F is assumed to be a Gaussian distribution with a mean of $\hat{f}(\mathbf{x})$ and a variance of $\hat{\sigma}^2(\mathbf{x})$. The expected improvement EI (x) is defined as the expected value of I (x) which is given by the following equation (Davins-Valldaura et al., 2017).

$$EI\ (\mathbf{x}) = \mathbb{E}_{F(\mathbf{x})}[I\ (\mathbf{x})] \tag{10}$$

The expected improvement can be expressed using the following equation.

$$EI\ (\mathbf{x}) = \begin{cases} \left[ f_{min} - \hat{f}(\mathbf{x})\Phi\left(\frac{f_{min}-\hat{f}(\mathbf{x})}{\sigma_k(\mathbf{x})}\right)\right] + \left[\sigma_k(\mathbf{x})\phi\left(\frac{f_{min}-\hat{f}(\mathbf{x})}{\sigma_k(\mathbf{x})}\right)\right] & if\ \sigma_k(\mathbf{x}) > 0 \\ 0 & if\ \sigma_k(\mathbf{x}) = 0 \end{cases} \tag{11}$$

Here, $\Phi$ and $\phi$ refer to the cumulative distribution and probability density functions of a Gaussian distribution. Maximizing the expected improvement results in the selection of the next parameter set for evaluating the function f. To optimize multiple objectives, the ParEGO algorithm first combines all objectives linearly to convert the problem into a mono-objective optimization problem. An iteration of the EGO method is then applied to this problem. During each iteration of the algorithm, a random weight vector $\lambda^j$ is generated using the following equation.

$$\lambda^j = \{\lambda_1^j, \lambda_2^j, ..., \lambda_n^j\}\ |\ \lambda_k^j \in [0,1] \quad \forall k = \{1,2,...,n\}\ \ and\ \ \sum_{k=1}^n \lambda_k^j = 1 \tag{12}$$

It is utilized by the Tchebycheff function $f_\lambda^j$, which linearly combines the objectives below.

$$f_\lambda^j(\mathbf{x}) = \max_{k=1,2,...,n} \left(\lambda_k^j f^j(\mathbf{x})\right) + \rho \sum_{k=1}^n \lambda_k^j f^j(\mathbf{x}) \tag{13}$$

The constant $\rho$ is set to a value of 0.05 (Knowles, 2006). The optimal points obtained from several mono-objective optimizations are then utilized to determine the Pareto Front. The qParEGO approach reduces the computational effort and can be particularly advantageous when time is a critical factor in conducting expensive evaluations in batches (Biswas et al., 2021).

### 4.6. PHV

This metric is used to evaluate the quality of a Pareto front approximation in multi-objective optimization. In this metric, the proportion of the HV enclosed by the true Pareto points is calculated for the resulting Pareto points (Aboutaleb et al., 2017).

$$PHV = \frac{HV(\text{ Resulted Pareto Points })}{HV(\text{ True Pareto Points })} \tag{14}$$

$PHV$ is a measure within the range of [0, 1], and in an ideal case, $PHV$ is equal to 1.

### 4.7. Filter-based Methods for Feature Selection

Filter-based methods are techniques that operate independently of the learning algorithm and use the inherent properties of data. Specifically, these methods utilize statistical correlations between a group of features and the target feature. The level of correlation between features and the target variable determines the significance of the target variable (Ottom, Chetty & Tran, 2013). So, filter-based methods can be a good option for feature selection when the number of data is low.

### 4.8. Generational Distance (GD)

The performance of an algorithm can be evaluated using the GD performance indicator, which calculates the distance from a solution to the Pareto front (Van Veldhuizen, 1999). Suppose the set of objective vectors, $A = \{a_1, a_2, a_3, \ldots, a_{|A|}\}$, is constructed from the points found by our algorithm, and the set of reference points (Pareto front) is denoted by $Z$ where $Z = \{z_1, z_2, z_3, \ldots, z_{|Z|}\}$. GD is then calculated in the following equation, where $d_i$ represents the Euclidean distance ($p = 2$) from $a_i$ to its nearest reference point in $Z$. In essence, this translates to the mean distance between any point in $A$ and its nearest point on the Pareto front.

$$GD(A) = \frac{1}{|A|} \left( \sum_{i=1}^{|A|} d_i^p \right)^{\frac{1}{p}}$$

### 4.9. Inverted Generational Distance (IGD)

Another performance indicator, IGD works by reversing the generational distance, and instead measures the distance from any point in $Z$ to its closest point in $A$ (Ishibuchi et al., 2015). The equation for IGD is given in the following where $\hat{d}_i$ represents the Euclidean distance ($p = 2$) from $z_i$ to its nearest reference point in $A$.

$$IGD(A) = \frac{1}{|Z|} \left( \sum_{i=1}^{|Z|} \hat{d}_i^p \right)^{1/p}$$

### 4.10. Proposed Multi-Objective Bayesian Optimization with Sequential Learning (MOBOSL)

In this section, a new data-driven multi-objective Bayesian optimization framework is presented. The proposed method consists of three main phases: Initialization, New Candidate Generation, and Model Updating.

### 4.10.1. Initialization

The method starts by setting certain input criteria, such as the maximum number of iterations, the HV threshold, and the hyperparameter search space, which are used to guide the optimization process. These input criteria are chosen based on the problem at hand and the desired optimization goals. Next, the data are scaled between 0 to 1. Then, the initial data is selected, and a Gaussian process (GP) is fitted to the initial data.

### 4.10.2. Generating the Next Candidate

In this step, the qParEGO acquisition function is used to select the candidate for the new sample point. Since our proposed approach will be illustrated using an already experimented dataset, this point is likely to be not included in this dataset. Therefore, we calculate its Euclidean distance from all existing points in the dataset to find the closest point. This closest point is then considered a new sample and is used to update the GP model. If the functional form of the data is known or the method is implemented in a full active learning mode, we can proceed with the datapoint returned by qParEGO directly.

### 4.10.3. Model Updating

The surrogate model (GP) is updated with the new candidate and steps one and two are repeated until the number of iterations is reached. The GP model is updated iteratively by selecting new points to evaluate based on a trade-off between exploration and exploitation. Specifically, the model updating process balances between exploring new regions of the search space, which may lead to a better solution, and exploiting the current knowledge to find the best solution. At the end of the process, the Pareto front of the resulting set of samples is determined. The Pareto front represents the set of non-dominated solutions that cannot be improved without sacrificing the improvement of another objective.

As depicted in Figure 7, the MOBOSL method follows an overall procedure of:

- The MOBOSL starts with setting input criteria, scaling the input data, selecting initial data far from the actual Pareto front, and fitting a Gaussian process to the data.
- The qParEGO acquisition function is used to generate new candidate sample points, and the Euclidean distance of each candidate point from all existing points in the dataset is calculated.
- The new candidate with the minimum distance is identified as a new sample and the optimization process is repeated until the maximum number of iterations or HV threshold is reached. The Pareto front is then determined as the set of non-dominated solutions.

Also, the pseudo-code of this algorithm is shown in Figure 8.

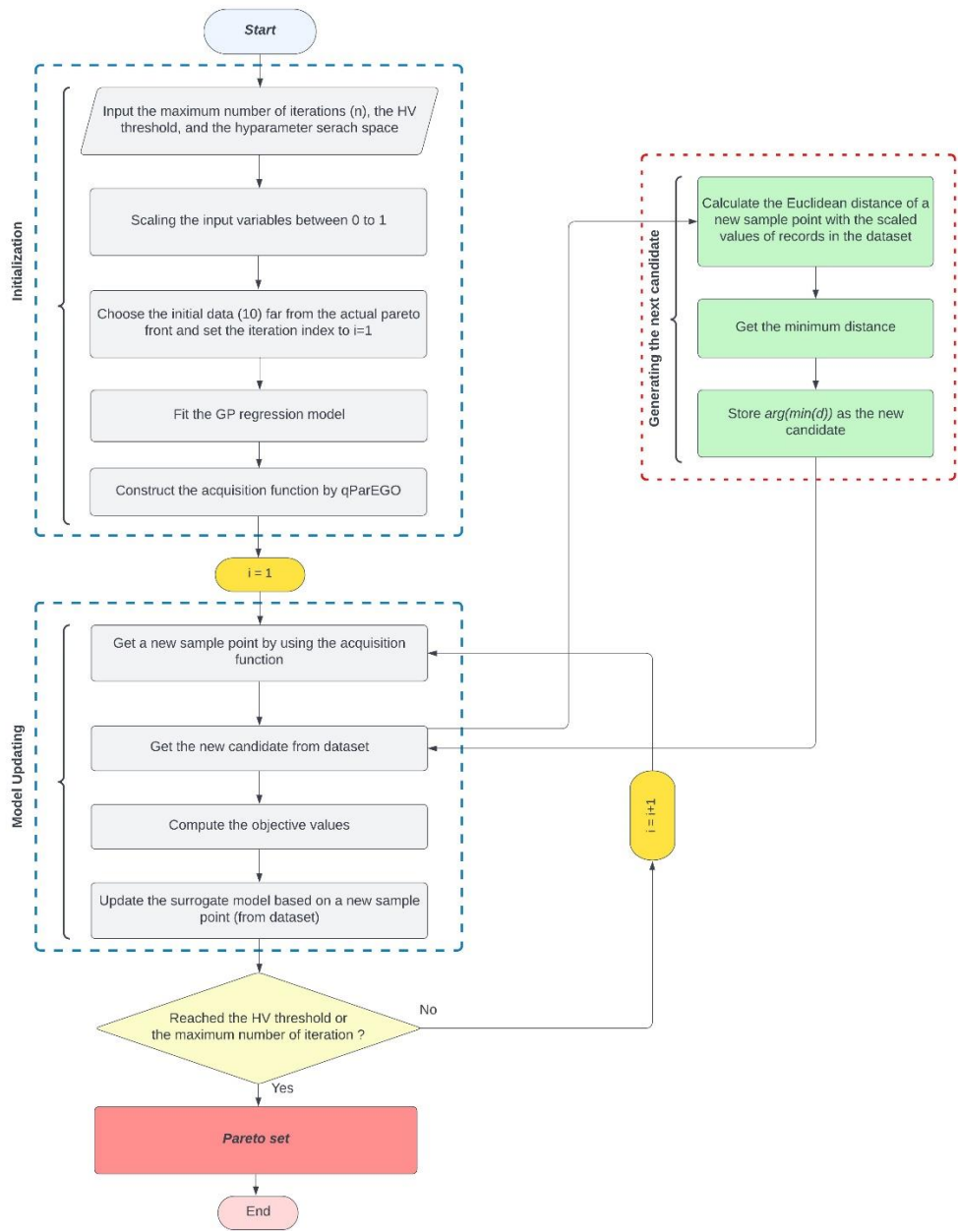

**Fig 7**: The MOBOSL flowchart

| Algorithm 1: A Novel Data-driven Multi-objective Bayesian Optimization |
|---|
| **Inputs**: Design space (dataset) $D$; let $l$ be the number of records in $D$, black-box objectives $f(\mathbf{x}) = (f_1(\mathbf{x}), f_2(\mathbf{x}), \ldots, f_m(\mathbf{x}))$; maximum number of iterations $n$; number of initial samples $k$; batch size $b$; the number of dimensions in the search space $num\text{-}dims$; the HV threshold $hv$ |
| **Output**: The set of points evaluated $\mathbf{X}$; the values of the objectives $\mathbf{Y}$; the set of non-dominated points $(\mathbf{X}, \mathbf{Y})$ in Pareto front |
| |
| *// Step 1: Initializing* |
| 01: Initialize $\mathbf{X}$ and $\mathbf{Y}$ by selecting points from the search space far from the actual Pareto front and evaluating $f(\mathbf{X})$ at $\mathbf{X}$ |
| 02: Scale the input variables between 0 to 1 |
| 03: Fit a Gaussian regression to the initial data $(\mathbf{X}, \mathbf{Y})$ |
| 04: *While* $i \leq n$ and $HV \leq hv$ **do** |
|       *// Step 2: Generating the next candidate* |
| 05:   Select a new sample point by $qParEGO(\mathbf{x_{new}})$ |
| 06:   *for* $j = 1$ to $l$ **do** |
|       *// Calculate Euclidean distance* |
| 07:       $d_m[j] = \parallel \mathbf{x_{new}} - D[j] \parallel$ |
|     *// Find the index of the minimum distance* |
| 08:    $min\text{-}index = argmin\,(d_m)$ |
|     *// Set x-next to the point at the min_index in D* |
| 09:    $\mathbf{x_{next}} = D[\min\_index]$ |
|     *// Step 3: Model update* |
| 10:    Evaluate $f(\mathbf{x})$ at $\mathbf{x_{next}}$ to obtain $\mathbf{y_{next}}$ |
| 11:    Update $\mathbf{X}$ and $\mathbf{Y}$ by adding $\mathbf{x_{next}}$ and $\mathbf{y_{next}}$ |
| 12:    Update the Gaussian regression and $qParEGO$ with $(\mathbf{x_{next}}, \mathbf{y_{next}})$ |
| 13: **Return** $\mathbf{X}$, $\mathbf{Y}$, and the Pareto front of $(\mathbf{X}, \mathbf{Y})$ |
| 14: *end* |

**Fig 8**: The MOBOSL Pseudo-code

## 4.11. APHV

The literature review section reveals the limitations of the existing criteria utilized for evaluating MOO. One critical flaw is the failure to consider the data utilization aspect, which is a significant factor in many fields, particularly in manufacturing where conducting additional experiments and gathering more data is often costly. Therefore, achieving optimal results while minimizing the number of records used is essential. To address this issue, a metric has been developed to cover both objectives. The APHV metric is based on the ratio of data utilization and PHV. Therefore, this method is especially useful in situations where resources are limited, and accurate and efficient solutions are required.

### 4.11.1. An Overview of the APHV Metric Formula

This criterion facilitates the comparison of MOO algorithms and considers that a lower usage of datapoints might be favored over a higher level of accuracy. The formula for the new criteria is defined in equation (11):

$$APHV = \alpha \times (1 - K) + \beta \times PHV \tag{11}$$

Where the variable "$K$" is defined as the ratio of data utilization. The data utilization is weighted by the parameter "$\alpha$", whereas the Pareto Front accuracy ('PHV') is weighted by the complementary parameter $(1 - \alpha)$ or "$\beta$". The metric ranges from zero to one, and higher values indicate better performance.

### 4.11.2. Deciding α and β for the APHV Metric

APHV is a combination of PHV and the number of used records. This metric takes a value between zero and one, where a value closer to one indicates a better Pareto front and less use of data points or manufacturing experiments. The output of this proposed metric is significantly affected by parameters $\alpha$ and $\beta$. These parameters are adjustable according to the specific problem and dataset at hand, which allows researchers and practitioners to optimize the metric to better align with the specific characteristics of the problem being addressed.

In this paper, to design the APHV metric, the values of $\alpha$ and $\beta$ are determined separately for the two scenarios. In the first scenario, when both objectives should be maximized, the $PHV$ with initial data turns out to be 0.55 $((1 - K) = 0.97)$, and with all the data $((1 - K) = 0)$ it is 1. We consider both of these situations to be equally undesirable and set $\alpha$ and $\beta$ in a way ($\alpha$=0.3 and $\beta$=0.70) so that the APHV will take the same value (APHV= 0.67) for both cases. On the other hand, in the second scenario, where one objective should be maximized and the other minimized, the PHV with initial data is 0.70 $((1 - K) = 0.97)$, and with all the data is 1. Just like the first scenario, we consider both of them as equally undesirable (APHV=0.75) by setting $\alpha$=0.2 and $\beta$=0.80. These values could be changed depending on the initial data points, underlying process, and model followed.

According to our setting, when the $PHV$ is 0, the Pareto front is not optimal, and the maximum value of the criterion is less than 0.3 and 0.2 (scenarios one and two, respectively), indicating poor quality of the front. In contrast, when the $PHV$ is 1, it means that the Pareto front is optimal, and the minimum value of the criterion is greater than 0.7 and 0.8, indicating a good quality of the front. As $K$ increases, the data utilization ratio increases, and the first part of the criterion decreases, which is consistent with the objective of minimizing $K$. Therefore, the APHV is an effective measure of the quality of the Pareto front and can guide the selection of the most appropriate solution among a set of non-dominated solutions. The adaptability of this criterion is noteworthy as it allows for the weights to be adjusted based on specific objectives and areas.

## 5. Results and Discussion

This section presents an analysis of the proposed MOBOSL model using various performance criteria, including the proposed APHV metric. First, multiple feature sets are selected using filter-based methods, physical insights, and experts' opinions. The ultimate objective is to obtain a high-quality Pareto front with a minimal number of input features and data points. Second, the primary parameters of the proposed model are defined. Subsequently, an evaluation of the model procedure, the quality of the resulting Pareto fronts, and a comparison of the model considering different feature sets are provided.

### 5.1. Determining Different Sets of Input Features

For selecting the different sets of features, we consider filter-based techniques, physical insights, and experts' opinions. Similar to Talapatera et al., we employ various sets of features from different perspectives to investigate the impact of different sets of features. By carefully selecting appropriate features, we could reduce the model's complexity and mitigate overfitting. Furthermore, the process of choosing the right feature subset also improves the model's accuracy.

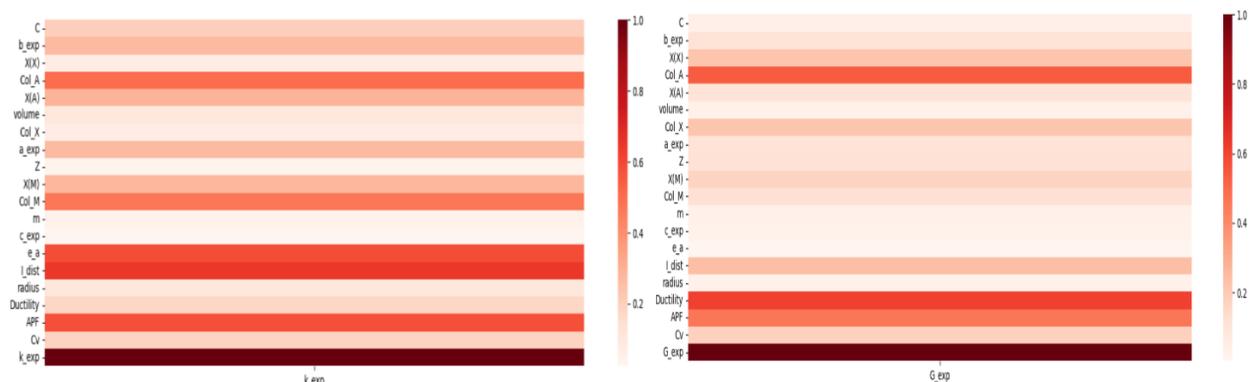

**Fig 9**: The Relative Importance of the Features for Both Objectives

The filter-based selection method employed in this study allowed us to determine the most influential features for achieving the two objectives. Utilizing a filter-based selection method, we examine the significance of features for both objectives and illustrate the results in Figure 9. Based on the relative importance of the features, Set 3 containing features with the highest relative importance (with a threshold of 0.60) is selected. Table 5 describes the selected sets.

Table 5: Different sets of selected features

| Set | Description | Features |
|---|---|---|
| Set1 | Physical insights and experts' opinion | 'Ductility', 'Type', 'APF', 'Col_A' |
| Set2 | Physical insights and experts' opinion | 'm', 'e_a', 'Z' |
| Set3 | Important Features | 'Ductility', 'I_dist' |

## 5.2. Defining the Primary Parameters of the MOBOSL Model

To apply the proposed model, it is necessary to establish certain initial parameter values. The key parameters of the algorithm are listed in Table 6.

Table 6: The primary parameters of the proposed MOBOSL approach

| Parameter | Description | Value(s) |
|---|---|---|
| $b$ | Batch size | 5 |
| $n$ | Maximum number of iterations | 150 and 90 |
| $NUM\_RESTARTS$ | The number of random restarts to perform when optimizing an acquisition function | 10 |
| $RAW\_SAMPLES$ | The number of random samples to generate from a given search space | 400 |
| $standard\ bounds$ | The lower and upper bounds of a search space (a tensor) | [0,1] |
| $MC\_SAMPLES$ | The number of Monte Carlo samples to use when estimating the expected improvement | 32 |
| $k$ | Number of initial samples | 10 |
| $l$ | Number of records in the dataset | 402 |
| $hv$ | The HV threshold | 0.97 |

Some of the parameters utilized in our proposed model are adapted based on Botorch (Balandat et al., 2020) and primarily used during the optimization phase. However, other parameters, such as $k$, $l$, and $n$, are determined based on the characteristics of the dataset and the specific cases presented in this paper. For instance, to mitigate manufacturing costs, we limit the amount of data utilized to obtain the Pareto front to no more than one-third and one-fourth of the available data for scenarios one (both objectives maximum) and two (one objective minimum and one objective maximum), respectively. Additionally, we aim to achieve a hypervolume of at least 0.97.

## 5.3. Analysis of the MOBOSL Model

In this section, we evaluate the MOBOSL model's performance using various sets and criteria in two different scenarios (both maximization and conflicting objectives). Initially, we examine the stability of the model by conducting 25 runs and analyzing the APHV values, which are presented in a box plot. Subsequently, we investigate the convergence behavior of the best, worst, and median runs of the model using the IGD indicator.

Furthermore, we provide an in-depth analysis of the proposed model's procedure for the median run. This analysis highlights how the model utilizes an initial sample, deliberately positioned far from the Pareto front, to optimize both objectives and obtain a Pareto front. Additionally, we include a comparison of the results based on different criteria to provide a comprehensive evaluation of the model's performance.

### 5.3.1. Max-Max Scenario

First, we evaluate the stability of the model in the three sets when the objectives are set to maximize. Figure 10 presents the box plot of 25 runs for each set. The APHV values range from 0.92 to 0.97 for set 1, which exhibits the most promising results. Set 3, on the other hand, shows the most stable results, with APHV values ranging from 0.78 to 0.81. In Set 2, where the actual Pareto front could not be obtained, the APHV criterion is the lowest but still exhibit a reasonable level of stability.

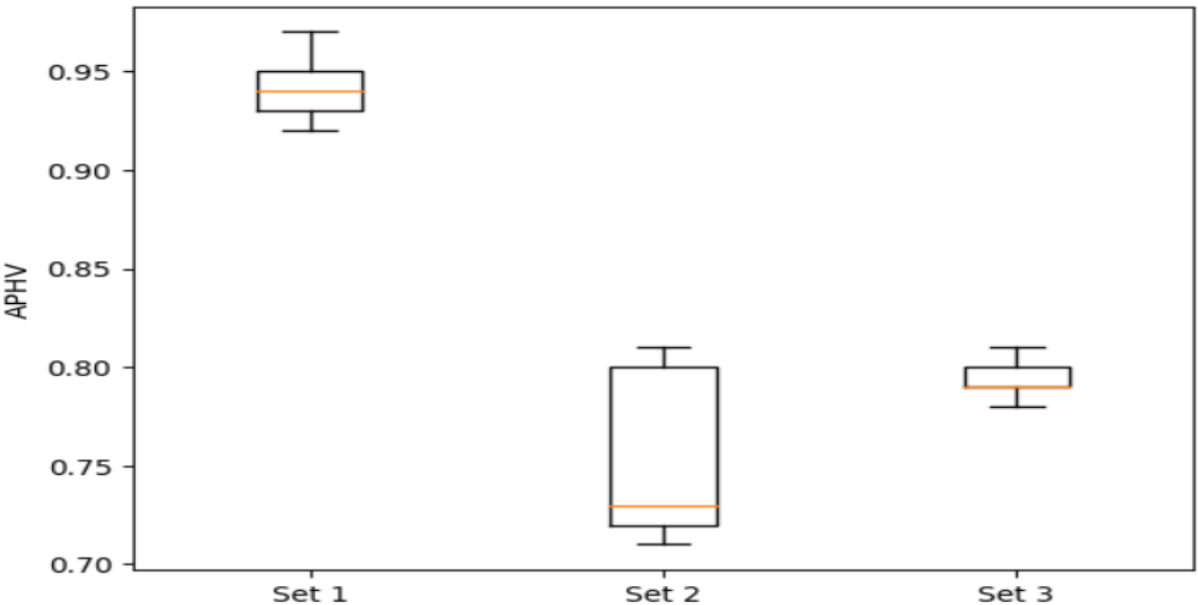

**Fig 10**: The stability of the model based on APHV in different runs (Max-Max)

Figure 11 shows the convergence towards the Pareto front for the best, median, and worst runs using the number of iterations as the x-axis and IGD criteria as the y-axis. The IGD criterion shows distinct patterns of convergence for different sets, with Set 1 converging faster than Set 2 and Set 3. In addition, it is noteworthy that the best run of Set 1 exhibits the highest convergence rate, with the Pareto front being achieved in approximately 40 iterations.

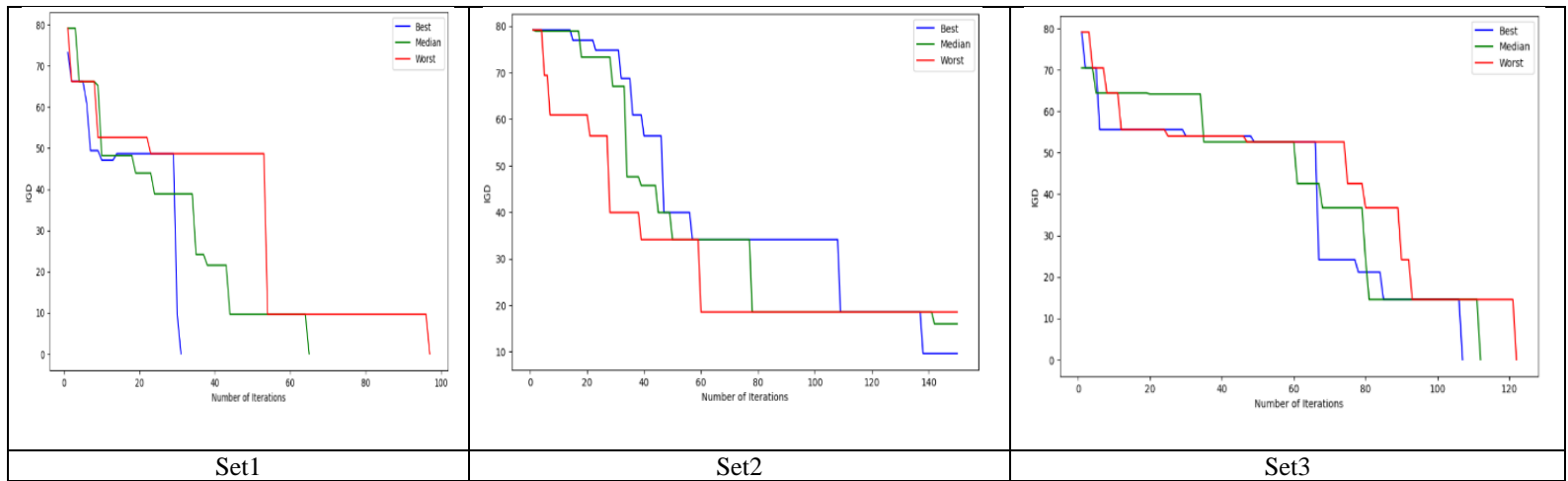

**Fig 11**: Convergence behavior of best, median, and worst run of the model based on IGD criterion (Max-Max)

Figure 12 illustrates the progression of the MOBOSL model. An evaluation of the sample status at different stages of the optimization process is presented in this figure, which shows the sample status after the first iteration, the sample status after half of the total iterations, as well as the final Pareto front.

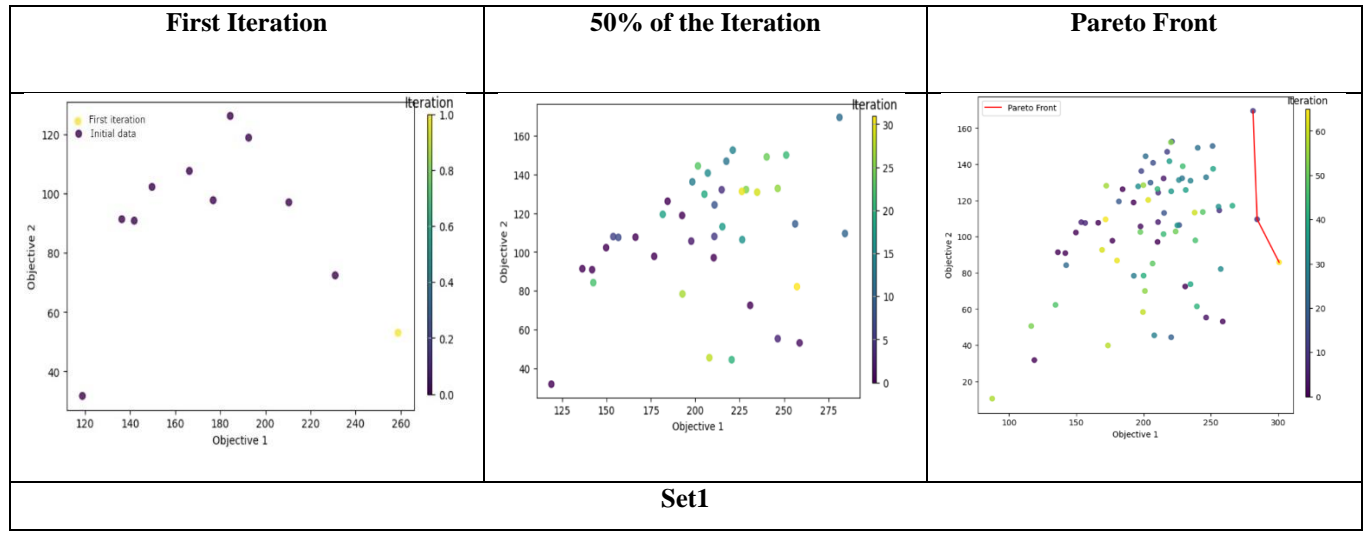

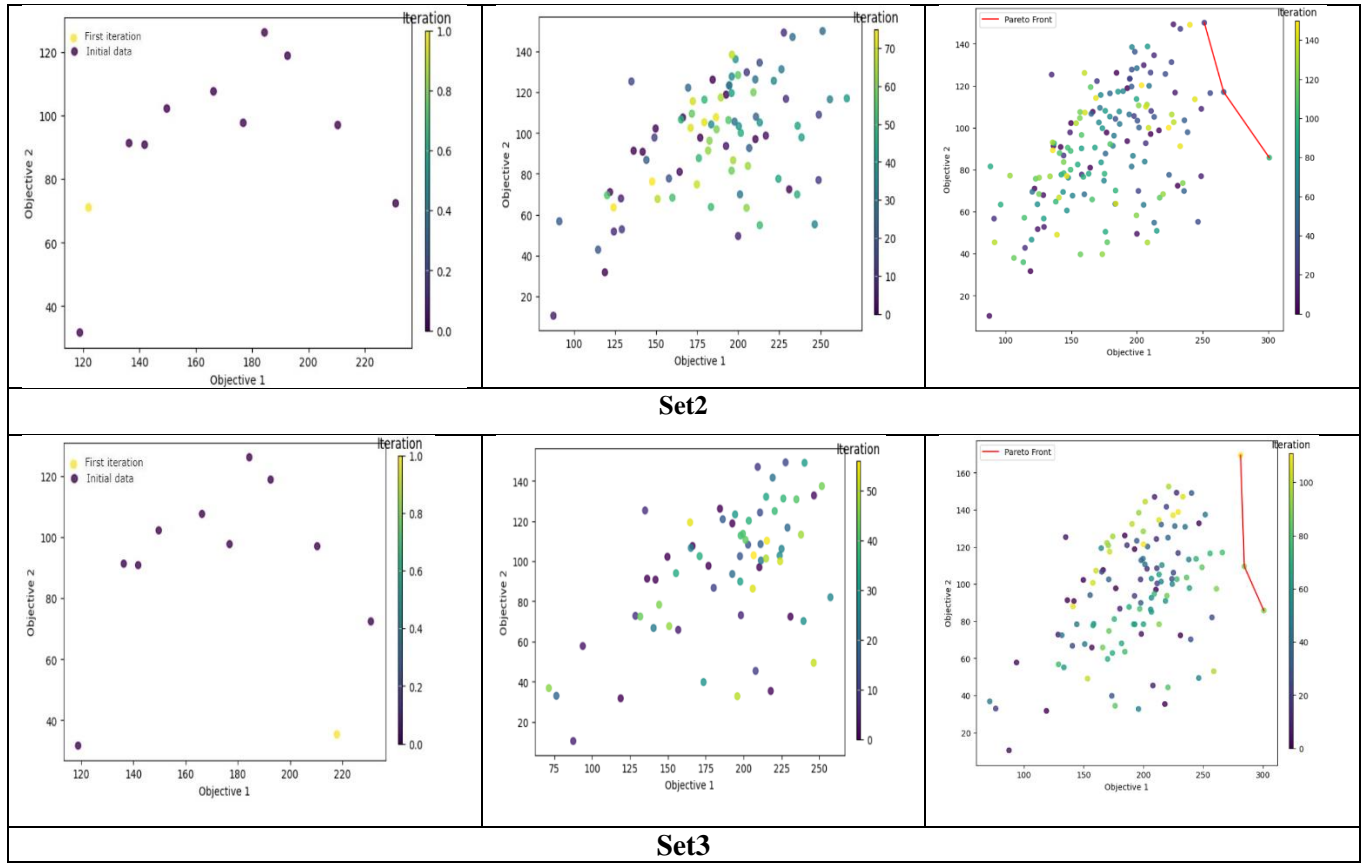

**Fig 12**: The procedure of the proposed model with different sets over iterations (Max-Max)

The above figure demonstrates that the initial random sample is not in proximity to the Pareto front. Nonetheless, the proposed model successfully optimizes both objectives and moves towards their maximum values, ultimately achieving Pareto fronts. It is worth noting that for both Sets 1 and 3, the model attains the true Pareto front. In contrast, for Set 2, the model reaches three Pareto points. To better comprehend the algorithm's performance, Figure 13 is presented, which shows the model's performance over different iterations within different criteria.

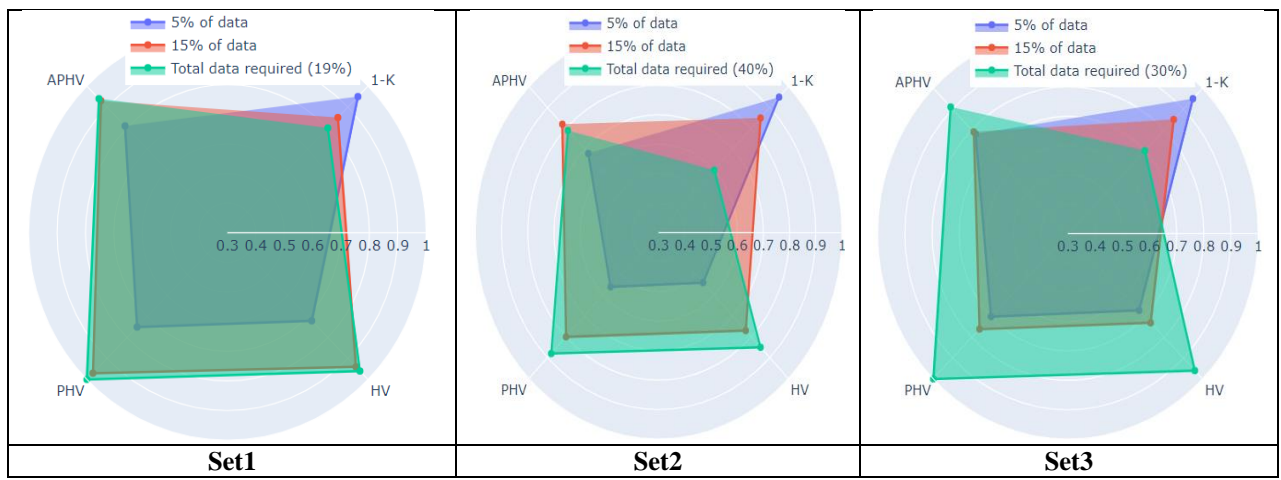

**Fig 13**: The comparison of the proposed model over different iterations with different criteria (Max-Max)

As demonstrated in the figure above, the lowest HV is obtained when Set 2 is considered. The new metric, APHV based on the PHV of 0.88 and the data utilization ratio of 0.40, yields a value of 0.79 for this model after reaching the maximum number of iterations (40%). The highest HV value is achieved when Set 1 and 3 are used with 19% and 30% of the data for modeling, respectively. APHV is also suggesting the same optimal combinations. Across all three spider plots, the new metric is observed to be effectively balancing the differences in PHV and data utilization ratio, thereby facilitating a more informed selection of the optimal model settings. Sometimes APHV provides different results than PHV. For example, considering the APHV criteria, using 50 iterations (15%) of Set 1 would be the second-best option despite having a lower PHV than Set 3 with 112 iterations (30%). Considering its ability to strike an optimal balance between a good Pareto front and minimal data consumption, APHV emerges as a more advantageous option.

### 5.3.2. Max-Min Scenario

The model stability is evaluated in three sets: one objective is set to maximize, and another to minimize. Figure 14 presents the box plot of 25 runs for each set, showing the distribution of APHV values.

Among the three sets, Set 3, where the actual Pareto front is achieved, exhibits the most promising results, with APHV values ranging from 0.96 to 0.99. This indicates that the model achieves high stability in approximating the Pareto front in Set 3. Set 1, on the other hand, shows the most stable results, with median APHV values reaching 0.91. In Set 2, the APHV criterion also exhibits a reasonable level of stability.

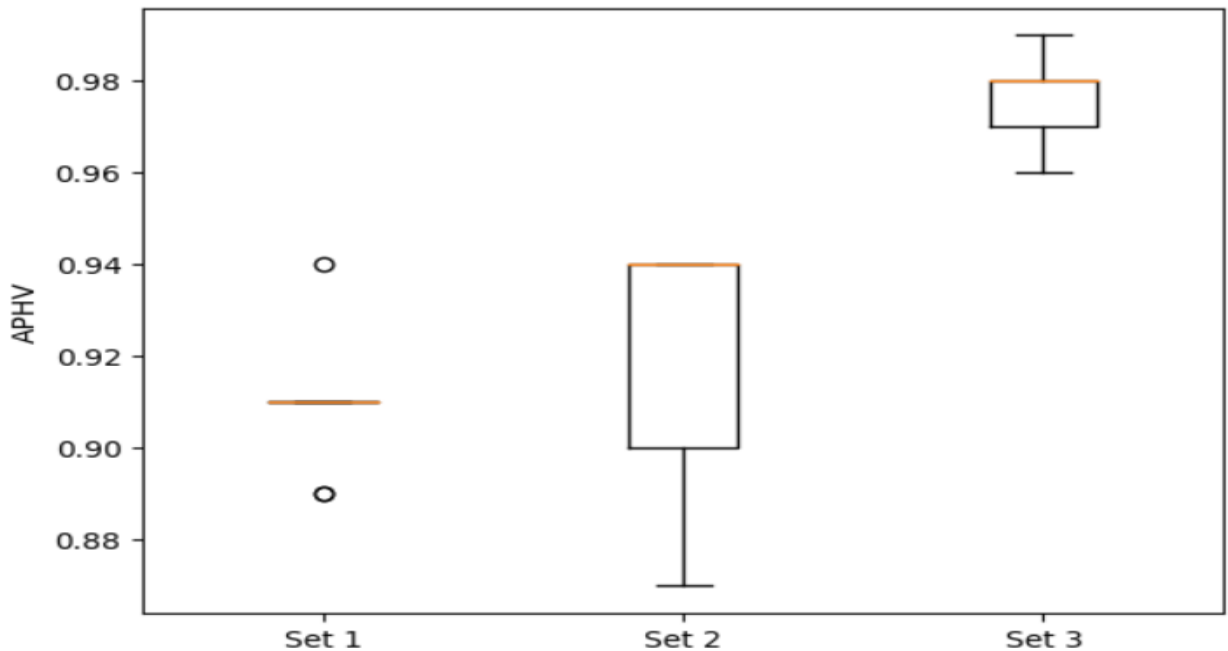

**Fig 14**: The stability of the model based on APHV in different runs (Max-Min)

Figure 15 shows the convergence rate of the best, median, and worst run of the model based on the APHV criterion.

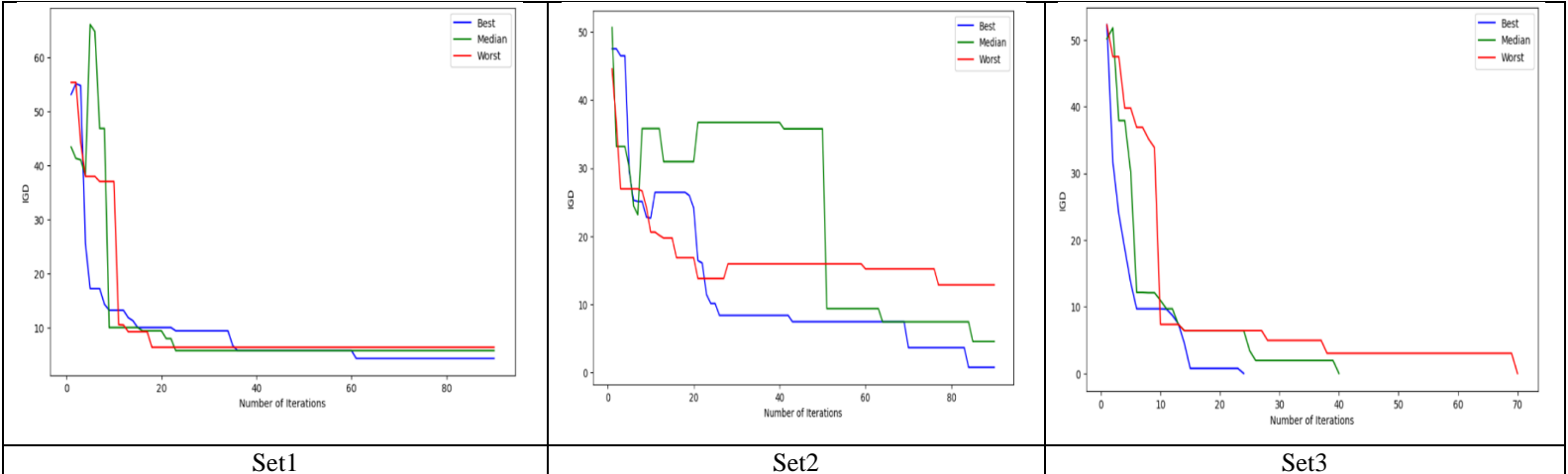

**Fig 15**: Convergence behavior of best, median and worst run of the model based on IGD criterion (Max_Min)

As shown in Figure 15, the model displays a high convergence rate in Set 1 and Set 3. Specifically, the best run in Set 3 has the fastest convergence rate while achieving the Pareto front in less than 30 iterations. The model with Set 2 has the slowest convergence rate. Overall, the models demonstrate favorable convergence rates across the different sets.

After analyzing the convergence rate, we compare our results with the findings of Talapatra et al. (2018). They reported the number of actual Pareto points obtained by their model at different numbers of iterations. It should be noted that the Pareto front in our case contains a total of ten points. We compare the results from our median runs in different sets with their best set, assuming a fixed number of initial data points set to 10.

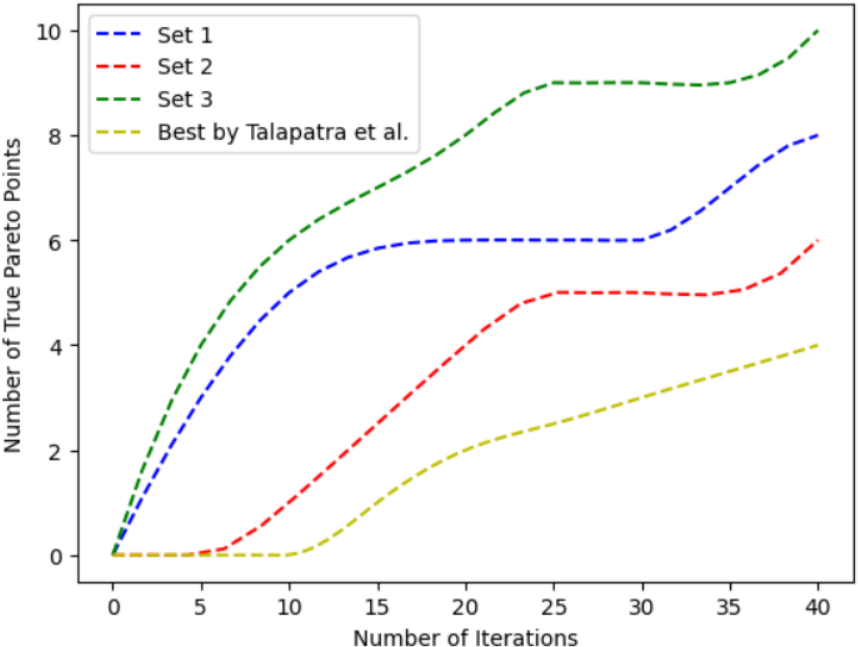

**Fig 16:** Number of true Pareto optimal points found by models in different sets

Based on Figure 16, it can be observed that for Set 3, our model achieves all the Pareto points in just 40 iterations. The model with Set 3 consistently obtains the actual Pareto front in all runs. Furthermore, our model with other sets also outperformed the best set reported by Talapatra et al. (2018), as evidenced by the superior results achieved in terms of the number of true Pareto points. The procedure utilized by the model to obtain their respective Pareto fronts is illustrated below.

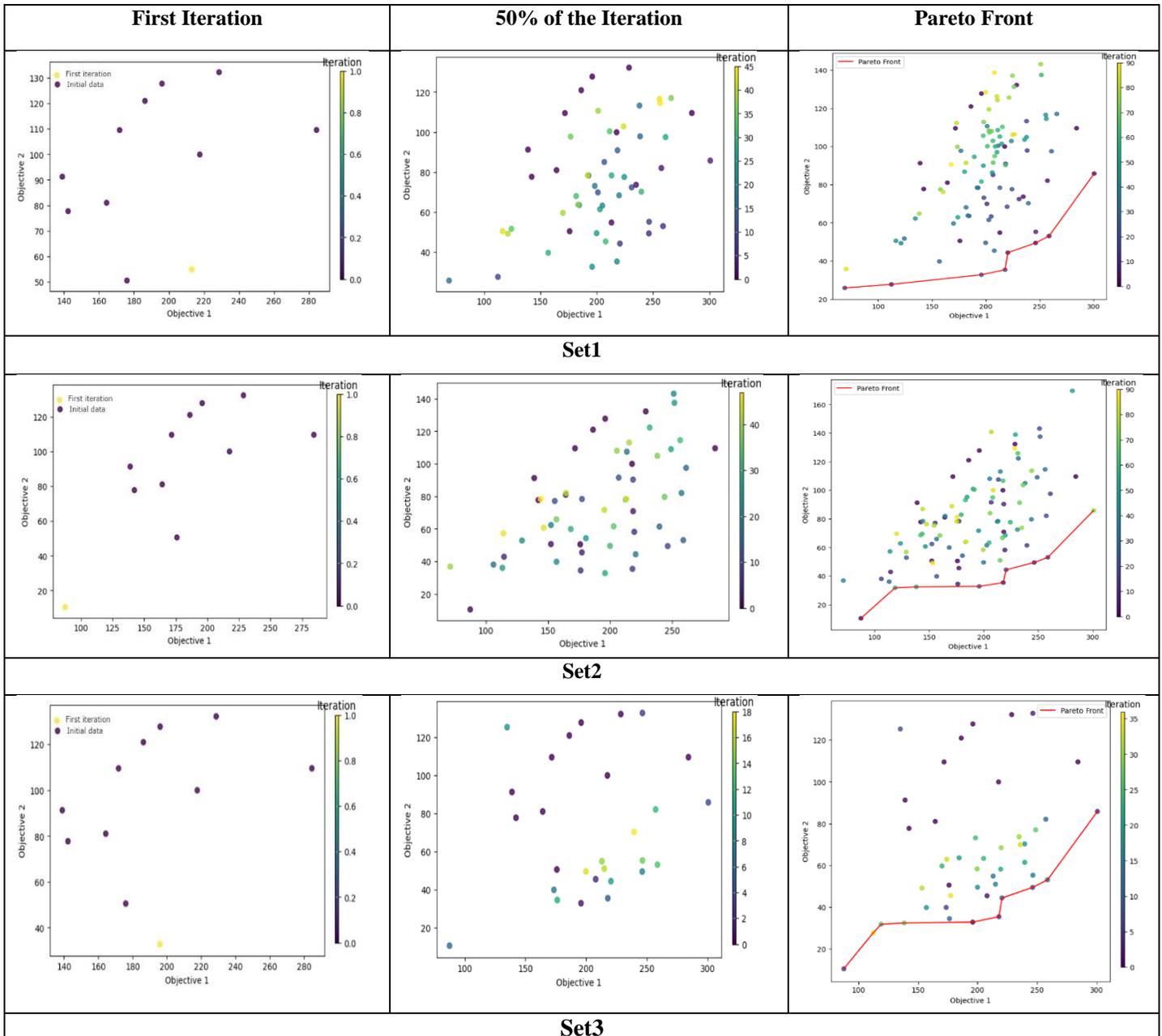

**Fig 17**: The procedure of the proposed model with different sets over iterations (Max-Min)

Similar to the max-max case, the depicted figure reveals that the initial random sample did not contain any true Pareto points. However, our proposed model successfully optimizes both objectives and progresses towards their maximum and minimum values, ultimately achieving Pareto fronts. Notably, for Sets 1 and 2, the model got 8 and 9 true Pareto points, respectively. In contrast, for Set 3, the model successfully reaches the actual Pareto front (PHV=1). To compare the performance of the models over different iterations, Figure 18 is presented.

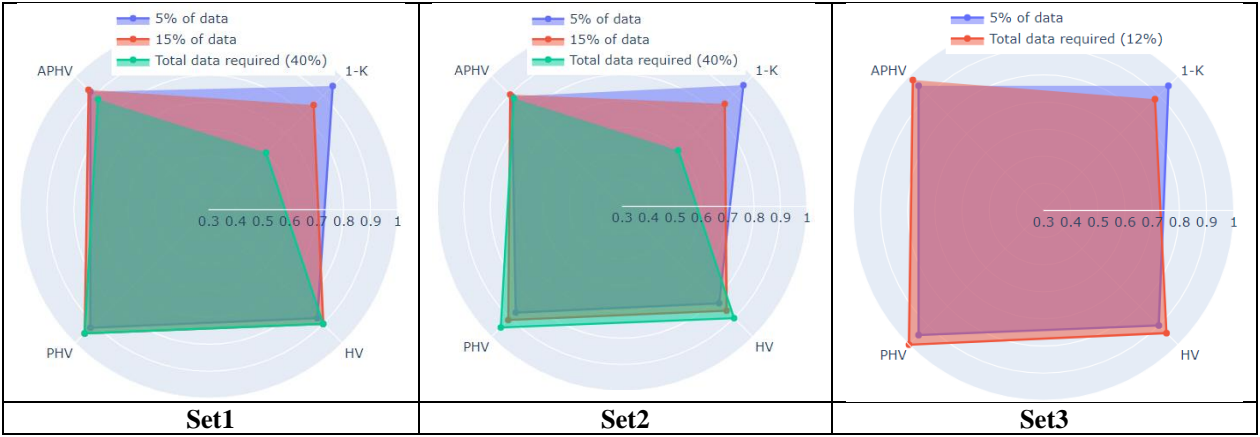

**Fig 18:** The comparison of the proposed model over different iterations with different criteria (Max-Min)

In Figure 18, a comparison of models is presented, considering different data ratios of 5%, 15% (if applicable), and total data required to reach the determined threshold. It can be observed that Set3, which provides the actual Pareto front, achieves the best performance with an APHV score of 0.98 when using 12% of the data. The model using 40% of the data in Set 1 has higher PHV and HV scores than the model with 15% data. However, based on APHV, the latter is identified as the better model with a score of 0.88 compared to 0.93. The APHV criterion proves to be a more reliable indicator of the model performance.

## 6. Conclusion

This paper proposes a data-driven multi-objective Bayesian optimization algorithm and illustrates its performance using a real-life manufacturing dataset. The model's superior performance in identifying optimal solutions with less data demonstrates its capability. Furthermore, a novel performance metric, APHV is also presented for evaluating multi-objective data-driven optimization approaches. This metric considers both the quality of the Pareto front and the cost of data generation, making it more practical for real-world applications.

The performance of the proposed model is evaluated in two different scenarios, namely maximization and conflicting objectives, using three distinct feature sets for optimization. The results show that Set 3 has the highest relative feature importance and achieves an average APHV of 0.97 in scenario 2, which involves conflicting objectives. Moreover, Set 3 can accurately identify the Pareto front in scenario 1. On the other hand, Set 1, although it fails to identify the entire Pareto front in scenario 2, displays promising results by identifying nine out of ten true Pareto points. It also reconstructs the Pareto front flawlessly in scenario 1 and produces more reliable results compared to the other sets. Set 2 performs poorly in comparison to Sets 1 and 3.

The proposed approach is highly beneficial in an active learning scenario where the objective is to identify the best possible combination of manufacturing process conditions or material properties to optimize one or more objectives. In most of these cases, material scientists or manufacturing engineers have little or no knowledge regarding the underlying function. In this problem setting, our model can reach the actual Pareto front using far less data than the traditional optimization algorithm. It promotes a sustainable manufacturing practice with less materials, man and machine hours. Future studies should concentrate on refining the current acquisition functions to strike a better balance between exploration and exploitation and to avoid trapping the model in a local optimum.

## Declaration of Competing Interests

The authors declare that they have no known competing financial interests or personal relationships that could have appeared to influence the work reported in this paper.

## Funding Information

This research has been generously supported by the United States Envirnmental Protection Agency (EPA) under the Pollution Prevention (P2) practices grant.

## Author Contributions

Conceptualization: Imtiaz Ahmed, Hamed Khosravi, Taofeeq Olajire; Methodology: Hamed Khosravi, Taofeeq Olajire, Imtiaz Ahmed; Formal analysis and investigation: Hamed Khosravi, Taofeeq Olajire; Writing - original draft preparation: Hamed Khosravi, Taofeeq Olajire, Ahmed Shoyeb Raihan; Writing - review and editing: Imtiaz Ahmed, Ahmed Shoyeb Raihan; Supervision: Imtiaz Ahmed.